\documentclass[conference]{IEEEtran}
\IEEEoverridecommandlockouts
\usepackage{cite}
\usepackage{amsmath,amssymb,amsfonts}
\usepackage{algorithmic}
\usepackage{graphicx}
\usepackage{textcomp}
\usepackage{xcolor}
\usepackage{enumitem}
\def\BibTeX{{\rm B\kern-.05em{\sc i\kern-.025em b}\kern-.08em
    T\kern-.1667em\lower.7ex\hbox{E}\kern-.125emX}}
\begin{document}

\title{Fast Convex Visual Foothold Adaptation for Quadrupedal Locomotion\\
}

\author{\IEEEauthorblockN{Shafeef Omar, Lorenzo Amatucci,
 Giulio Turrisi, Victor Barasuol, Claudio Semini}
\IEEEauthorblockA{Dynamic Legged Systems (DLS) Lab\\
Istituto Italiano di Tecnologia (IIT)\\
Genova, Italy \\
firstname.lastname@iit.it
}
}

\maketitle

\def\HyqReal{\centering
    \includegraphics[width=0.22\textwidth]{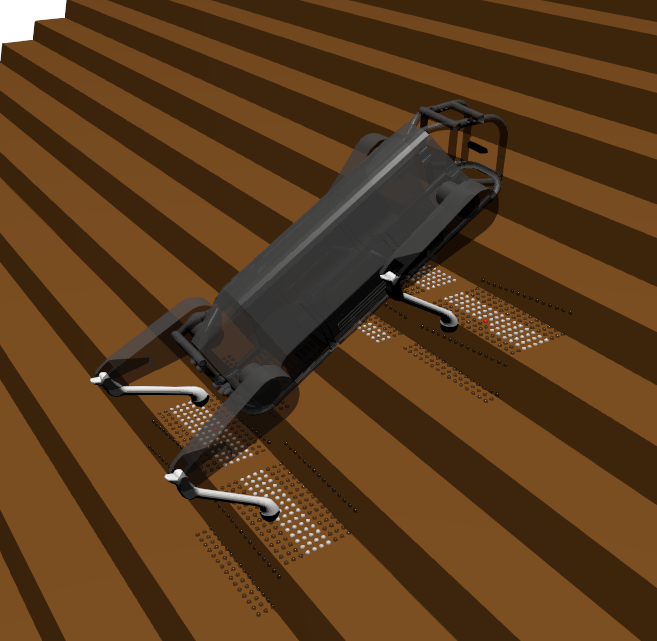}
    \hskip 0.3em
    \includegraphics[width=0.2135\textwidth]{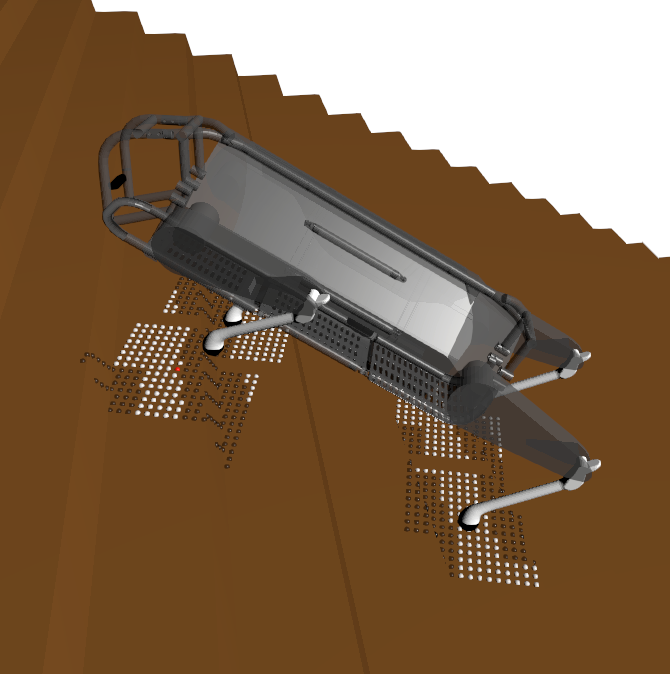} 
}

\def\Unet{\centering
    \includegraphics[width=1.\textwidth]{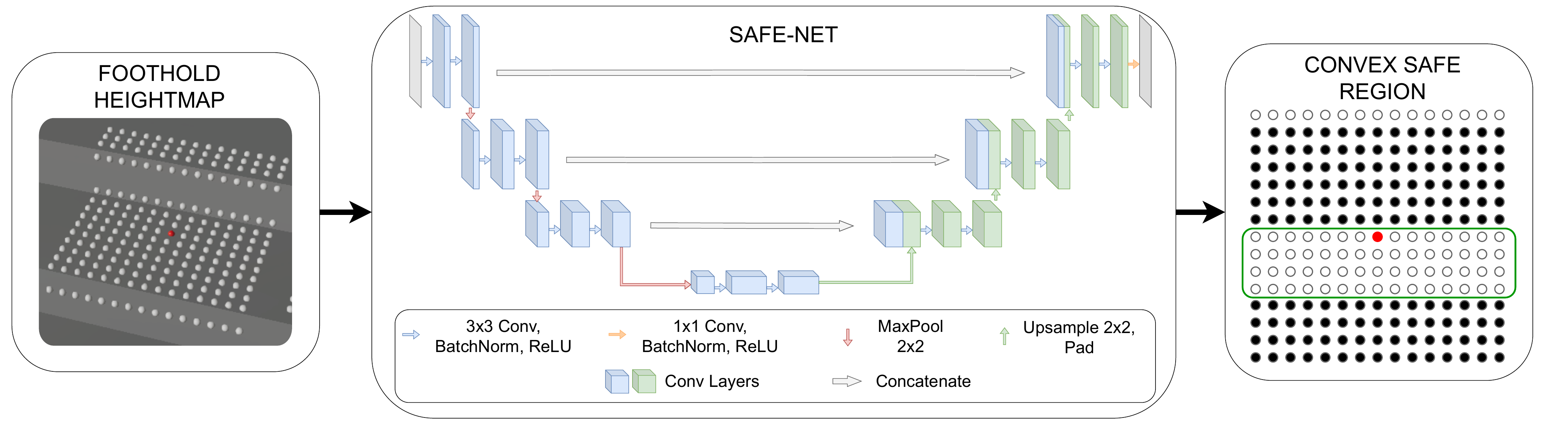}
}

\def\simulation{
\includegraphics[width=0.49\textwidth]{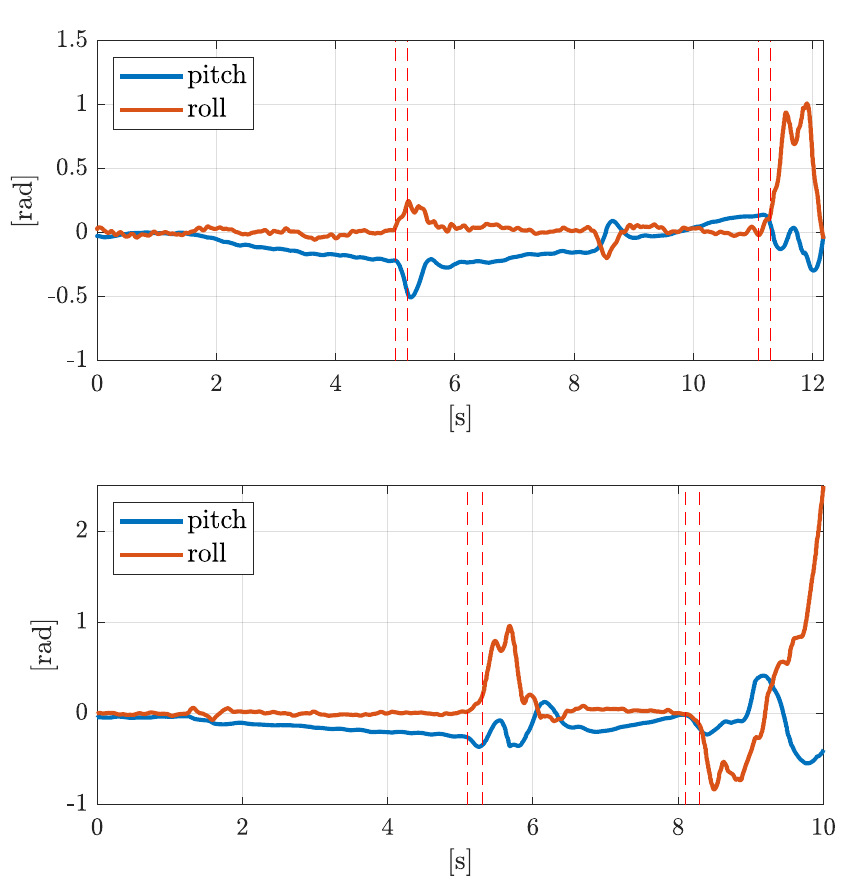}
}

\begin{abstract}
This extended abstract provides a short introduction
on our recently developed perception-based controller for quadrupedal locomotion. Compared to our previous approach based on Visual Foothold Adaptation (VFA) and Model Predictive Control (MPC), our new framework combines a fast approximation of the safe foothold regions based on Neural Network regression, followed by a convex decomposition routine in order to generate safe landing areas where the controller can freely optimize the footholds location. The aforementioned framework, which combines prediction, convex decomposition, and MPC solution, is tested in simulation on our 140kg hydraulic quadruped robot (HyQReal).
\end{abstract}


\section{Introduction}
Legged Locomotion is a challenging problem when considering both agile motions and rough terrains. In literature, many works have tackled the first problem, developing controllers that are able to plan complicated motions given the dynamical model of the robot, which can be based on a simplified assumption \cite{ba}, or on more accurate descriptions of the system such as the Single Rigid Body Dynamics (SRBD) \cite{b8}. Still, vision-based locomotion, required for traversing complex scenarios, can be difficult to tackle accurately. First, it brings the necessity of accurately describing the environments, and second, it constrains the possible robot foot position in order to avoid stepping on unsafe locations (e.g. edges). In this last regard, many attempts have been made considering a fixed and safe foot placement for the robot \cite{b12,b13}, but this assumption ultimately hinders the final performance of the controller. In fact, the benefit of jointly optimizing for the best
foothold location and robot motion have been demonstrated in \cite{b10,b11}. Recently, in \cite{b9}, this idea was augmented with a plane segmentation technique in order to generate terrain-aware convex foothold constraints. Similarly, in this extended abstract, we enhance our visual foothold adaptation technique \cite{b1} in order to generate safe landing areas where the controller
can freely optimize for the best footholds locations. Additionally to \cite{b9}, our method, thanks to our fast evaluation criteria (Sects.~\ref{sec:Safe_Foothold_Evaluation_Criteria},~\ref{sec:Feasibility_Net}), is able to automatically cope with kinematic constraints, leg and swing foothold collisions, without the necessity of any additional consideration inside the MPC scheme.
\begin{figure}[!h]
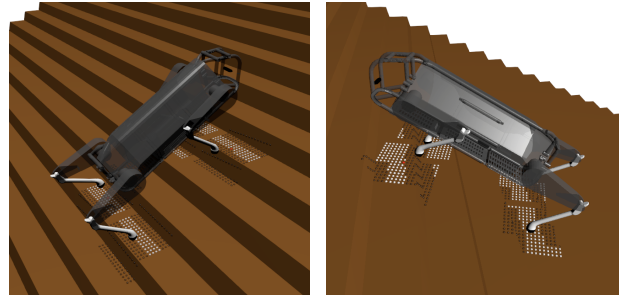

\HyqReal
\caption{IIT's HyQReal quadruped robot climbing stairs in RaiSim \cite{b3}.}
\label{fig:hyqreal_raisim}
\vspace{-15pt}
\end{figure}

\section{The Proposed Approach}

\begin{figure*}[!h]
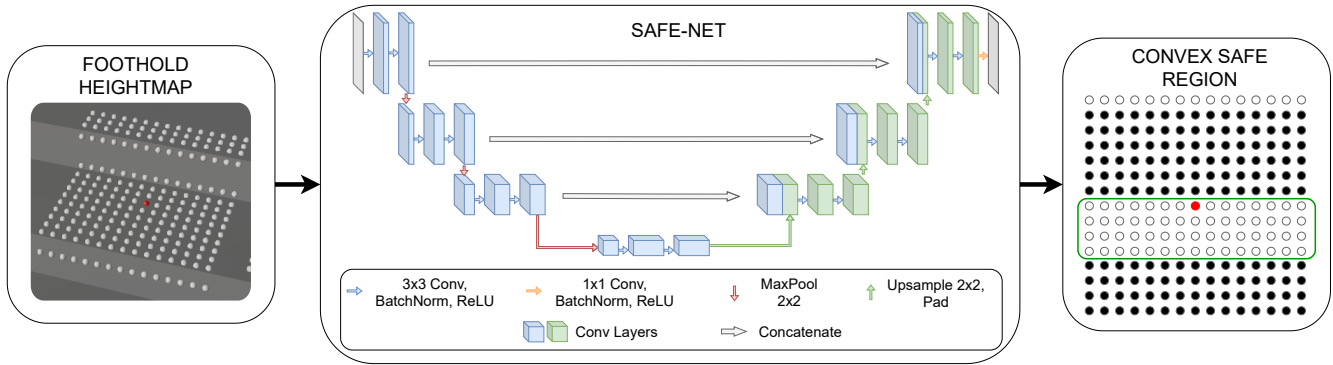

\Unet
\caption{Block diagram of the proposed framework. Starting from the left side, a foothold heightmap is acquired from sensor data and then processed by the Safety Net for fast foothold evaluation. In the end, a convex region is extracted from the output of the network.}
\label{fig:unet}
\vspace{-15pt}
\end{figure*}

\subsection{Safe Foothold Evaluation Criteria}
\label{sec:Safe_Foothold_Evaluation_Criteria}
Given a \textit{nominal} foothold location, we define as foothold heightmap a squared and discrete representation of the
terrain where each pixel is described by the height of a certain
area. On this local area representation, we run different criteria in order to evaluate the safety of each possible landing location, analyzing:
\begin{enumerate}[label=\alph*)]
\item Kinematic reachability: if a foothold is outside the workspace of
the robot leg, the pixel is discarded.
\item Terrain Roughness: for each heightmap pixel, we
compute the height difference relative to its neighborhood. If the foothold together with a \textit{minimum} number of neighborhoods are above a certain height threshold, the pixel is discarded.
\item Frontal Collision: for each pixel, we evaluate potential foot frontal collisions along the corresponding swing trajectory. Locations that bring possible collisions are discarded.
\item Leg Collision: Similar to frontal collisions, we
evaluate the intersection between the terrain and both leg
limbs throughout the whole step cycle (i.e., stance and swing
phases).
\end{enumerate}
After the evaluation of the foothold heightmap, we obtain a representation of the possible safe locations represented by non-convex and possibly disjointed regions. Previously in \cite{b1}, we choose as the optimal foothold the safe point \textit{nearest} to the nominal stepping location, but this heuristic neglects dynamics. Hence, from an optimization perspective, we aim to provide to the controller a convex area inside the obtained safe region where it can autonomously choose the best foothold giving its own cost function. 

\subsection{SaFE-Net and Dataset Collection}
\label{sec:Feasibility_Net}

The aforementioned criteria can be hard to run in real-time. In fact, these evaluations should be run for each foot of the quadruped and can take approximately 40~ms in total to be computed on an Intel Core i7-8700k. In order to speed up the process, we performed a Neural Network based regression that mimics the foothold heightmap evaluation (SAFE Net - SAfe Foothold Evaluation Network). We modified U-net \cite{b2}, a Convolution Neural Network which yields fast (around 1.5~ms of computation time for each leg) and precise segmentation of the foothold
heightmap. 
Generating a large number of training examples can be time-demanding since they depend on the actual state of the robot and on the terrain morphology. For this reason, we collected data from the visual foothold heuristic criteria (Sect.~\ref{sec:Safe_Foothold_Evaluation_Criteria}) in RaiSim \cite{b3}, a fast and parallelizable simulator that generates variegated rough terrains on demand (Fig.~\ref{fig:hyqreal_raisim}).

\subsection{Regions Decomposition and Model Predictive Control}
\label{sec:Regions_Decomposition}
The output of SaFE-Net cannot be used directly inside an optimal quadratic control problem, as explained in Sect.~\ref{sec:Safe_Foothold_Evaluation_Criteria}. For this reason, we perform a decomposition of the safety map in order to generate multiple convex regions using the library in \cite{b4}, enabling the possibility to add simpler constraints inside our optimization problem (see {Fig.}~\ref{fig:unet}). Deciding which of these regions to use can be solved by a demanding Mixed-Integer Program, but in our case, we choose as a candidate the region nearest to the nominal foothold location. Finally, the MPC formulation is similar to the one in \cite{b5}, which uses an SRBD model that neglects the inertia of the legs. We compensate for this assumption by applying an extra torque as in \cite{b1}.
State variables and control inputs are augmented to include future foothold optimization \cite{b6}.

\section{Results and Conclusion}
\begin{figure}[!h]
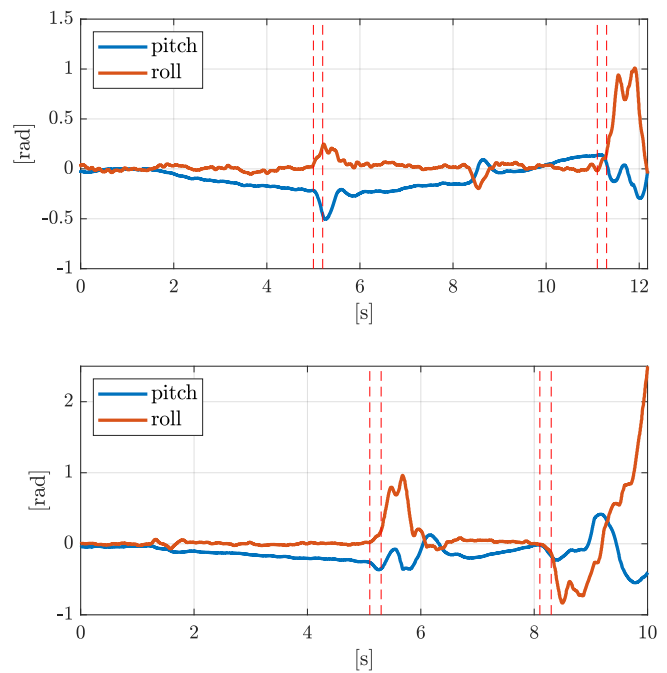

\simulation
\caption{Simulation results showing the pitch and roll angles of the robot while climbing the stairs scenario with the proposed approach (top) and with our previous method \cite{b1} (bottom). The application of the disturbance is indicated by a vertical dashed line.}
\label{fig:simulation}
\end{figure}

We present dynamic simulations on our hydraulic quadruped robot HyQReal. In the simulations (see the accompanying video\footnote{https://www.youtube.com/watch?v=OlmrsYRIwds}), the robot must walk forward while climbing some stairs and performing a trotting gait. During the motion, strong disturbances (700N for 0.2 second in the lateral direction) are added to the CoM of the robot to perturbate its dynamics.  In the first simulation, the controller does not have the ability to choose the optimal foothold location, which is provided heuristically from the safety map. Viceversa, in the second simulation, a safe convex region is passed to the controller. As shown in the video, only in the second case the robot is able to climb the stairs since a better (nevertheless safe) foothold is automatically chosen by the MPC to stabilize the system (Fig. \ref{fig:simulation}). 
Our scheme runs at a frequency of
150 Hz, and thus it is suitable for real-time implementation
on a physical robot. Future work will focus on testing the proposed framework on the real platform.

\end{document}